\documentclass{article}
\usepackage{amsmath,amssymb} % define this before the line numbering.

\PassOptionsToPackage{numbers, compress}{natbib}
\usepackage[preprint]{neurips_2020}

\usepackage[utf8]{inputenc} % allow utf-8 input
\usepackage[T1]{fontenc}    % use 8-bit T1 fonts
\usepackage{hyperref}       % hyperlinks
\usepackage{url}            % simple URL typesetting
\usepackage{booktabs}       % professional-quality tables
\usepackage{amsfonts}       % blackboard math symbols
\usepackage{nicefrac}       % compact symbols for 1/2, etc.
\usepackage{microtype}      % microtypography

\usepackage{subcaption}
\usepackage{floatrow}
\usepackage{mathtools}

\usepackage{array}
\usepackage{boldline , multirow}
\newcommand{\PreserveBackslash}[1]{\let\temp=\\#1\let\\=\temp}
\newcolumntype{C}[1]{>{\PreserveBackslash\centering}p{#1}}
\newcolumntype{R}[1]{>{\PreserveBackslash\raggedleft}p{#1}}
\newcolumntype{L}[1]{>{\PreserveBackslash\raggedright}p{#1}}
\newcolumntype{M}[1]{>{\centering\arraybackslash}m{#1}}
\usepackage{xcolor}
\usepackage{comment}
\title{B-SCST: Bayesian Self-Critical Sequence Training for Image Captioning}

\author{%
	Shashank Bujimalla\thanks{Equal Contribution} \\
	\normalsize{Intel Corporation}\\
	\texttt{\small shashankbvs@gmail.com}\\	
	\and
	\textbf{Mahesh Subedar}\footnotemark[1] \\
	\normalsize{Intel Labs}\\
	\texttt{\small mahesh.subedar@intel.com}\\	
	\and  
	\textbf{Omesh Tickoo} \\
	\normalsize{Intel Labs}\\
	\texttt{\small omesh.tickoo@intel.com}\\
}
\begin{document}

\maketitle

\begin{abstract}
Bayesian deep neural networks (DNNs) can provide a mathematically grounded framework to quantify uncertainty in predictions from image captioning models. We propose a Bayesian variant of policy-gradient based reinforcement learning training technique for image captioning models to directly optimize non-differentiable image captioning quality metrics such as CIDEr-D. We extend the well-known Self-Critical Sequence Training (SCST) approach for image captioning models by incorporating Bayesian inference, and refer to it as B-SCST. The “baseline” for the policy-gradients in B-SCST is generated by averaging predictive quality metrics (CIDEr-D) of the captions drawn from the distribution obtained using a Bayesian DNN model.  We infer this predictive distribution using Monte Carlo (MC) dropout approximate variational inference. %, which is one of the standard ways to
We show that B-SCST improves CIDEr-D scores on Flickr30k, MS COCO and VizWiz image captioning datasets, compared to the SCST approach. We also provide a study of uncertainty quantification for the predicted captions, and demonstrate that it correlates well with the CIDEr-D scores. To our knowledge, this is the first such analysis, and it can improve the interpretability of image captioning model outputs, which is critical for practical applications.
\end{abstract}
\vspace{-2pt}
\section{Introduction}
\vspace{-2pt}
Deep neural network (DNN) based image captioning approaches generate natural language descriptions of an image by transforming the image features into a sequence of output words from a predefined vocabulary. State-of-the-art image captioning models~\cite{anderson2018bottom,cornia2019m,huang2019attention,zhou2019grounded,xu2015show} use encoder-decoder architecture~\cite{vinyals2015show,karpathy2015deep,xu2015show,anderson2018bottom}, and follow a two-step training process.
In the first step, cross-entropy loss is optimized to generate a caption with words in the same order as the ground-truth caption. In the second step, policy-gradient based reinforcement learning (RL)~\cite{rennie2017self} is used to minimize the negative expected value of the generated caption quality metric scores (typically CIDEr-D)~\cite{papineni2002bleu,denkowski2014meteor,vedantam2015cider,anderson2016spice}. %However, the quality of image captioning is evaluated using Natural Language Processing (NLP) metric scores such as BLEU~\cite{papineni2002bleu}, METEOR~\cite{denkowski2014meteor}, CIDEr-D~\cite{vedantam2015cider} and SPICE~\cite{anderson2016spice}. These metrics are non-differentiable and cannot be directly maximized by this cross-entropy loss optimization algorithm. So, in the second step, policy-gradient based reinforcement learning (RL) is used to minimize the negative expected value of the caption scores~\cite{rennie2017self}.

%The encoder takes the image features, which are obtained using a pretrained CNN or Faster R-CNN model~\cite{ren2015faster}, and transforms them into latent space using techniques such as attention mechanism~\cite{xu2015show,anderson2018bottom}. The decoder, which usually includes RNNs or LSTMs, takes these latent features, and generates a sequence of words to form an image caption.The training of these state-of-the-art models typically follows a two-step process.

%In the first step, cross-entropy loss is optimized to generate the captions with words in the same order as the ground-truth captions. However, the quality of image captioning is evaluated using Natural Language Processing (NLP) metric scores such as BLEU~\cite{papineni2002bleu}, METEOR~\cite{denkowski2014meteor}, CIDEr-D~\cite{vedantam2015cider} and SPICE~\cite{anderson2016spice}. These metrics are non-differentiable and cannot be directly maximized by this cross-entropy loss optimization algorithm. So, in the second step, policy-gradient based reinforcement learning (RL) is used to minimize the negative expected value of the caption scores~\cite{rennie2017self}.

Several recent works have shown that using a bias correction, i.e., a learned “baseline”, to normalize the RL rewards reduces the variance in policy gradients, and is effective during training. In Self-Critical Sequence Training (SCST)~\cite{rennie2017self}, CIDEr-D metric is directly optimized. The model chooses word with the highest SoftMax probability at each timestep and generates a greedy caption, i.e., the caption generated using the inference algorithm. The CIDEr-D score of this greedy caption is used as the “baseline” to increase the probability of generating captions that have higher score than the "baseline" while decreasing  it for captions that have lower score.%, thus optimizing the CIDEr-D metric directly.

\begin{figure}[t]
\centering
\includegraphics[width=0.90\linewidth]{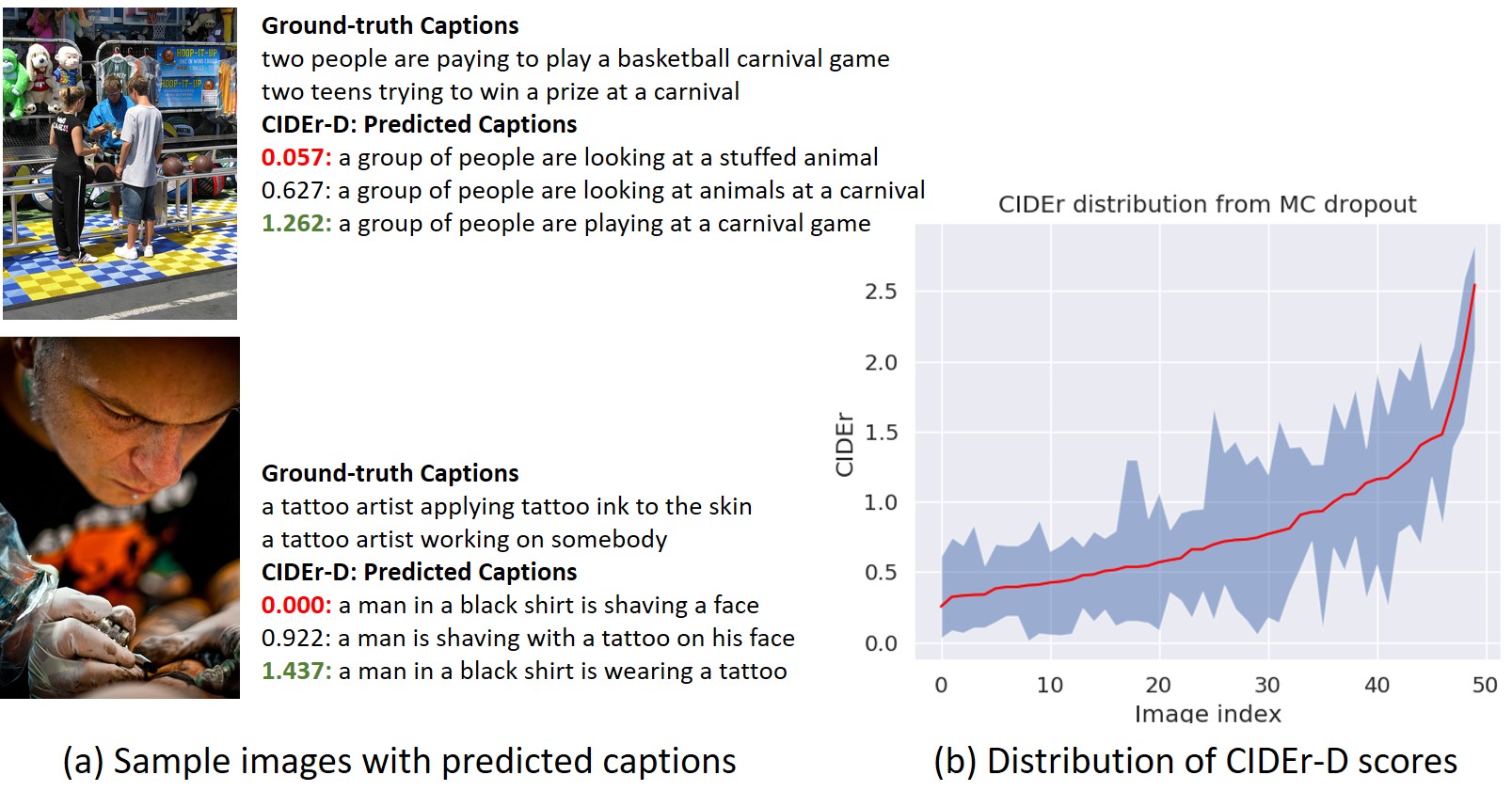}
\caption{\small (a) Sample images and few of their ground-truth captions from Flickr30k dataset. Also shown are the greedy captions that are predicted using a trained model with MC Dropout forward passes, along with their corresponding CIDEr-D scores (red is low, green is high). (b) Distribution (min-max) of CIDEr-D scores obtained from MC Dropout forward passes for 50 randomly selected images from Flickr30k dataset, sorted by their predictive mean CIDEr-D scores~(shown in red).}
\label{fig:example}
\end{figure}

%Although deep neural networks (DNNs) provide state-of-the-art results for a multitude of applications, they have been shown to fail\cite{gal2016uncertainty} in the case of noisy or out-of-distribution data leading to overly confident SoftMax probability scores.
Image captioning DNN models can generate incorrect description of a given image, hence it is important to study the inherent ambiguity or uncertainty estimates of the generated captions.
Bayesian DNNs~\cite{blundell2015weight,gal2016uncertainty} provide a principled way to gain insight into the data and capture reliable uncertainty estimates, leading to interpretable models.
%Bayesian DNNs~\cite{blundell2015weight,gal2016uncertainty} provide a principled way to gain insight into the data and model uncertainty, and capture reliable uncertainty estimates leading to interpretable models.
In this work, we use MC dropout approximate inference~\cite{gal2016dropout} to study the correlation between uncertainty estimates and the CIDEr-D score of the generated captions. This analysis is critical to enable practical image captioning applications that require interpretability of the model outputs.

In Figure~\ref{fig:example}~(a), we show two sample images from Flickr30k~\cite{plummer2015flickr30k} dataset along with few of their ground truth captions. We also show the greedy captions that are generated from a trained model using MC dropout forward passes, and their corresponding CIDEr-D scores.
%MC Dropout provides a way to sample captions from the posterior distribution, and an average predictive score of these captions can be better estimate of the “baseline” score.
%We also show later in Section~\ref{sec:results} (Figure~\ref{fig:uncert-vs-cider-flickr}) that captions with lower CIDEr-D scores have higher uncertainty indicating better correlation, although their SoftMax probabilities are relatively uniform and high.
%SCST approach uses a single greedy caption, which has highest SoftMax probability, as the “baseline” during training to improve or suppress the probability of sampled captions, and may pick any one of these high variance CIDEr-D scores as the “baseline” score.
%The greedy predictions may endup using one of the predicted captions, but an average of distribution of predicted captions will be better representation of baseline caption from the model.
%\begin{comment}
%Since SCST approach uses a single greedy caption, it may pick any one of these high variance CIDEr-D scores, as the "baseline" during training.
During training, SCST approach uses CIDEr-D score of a single greedy caption, which is obtained from standard DNN inference algorithm, as the "baseline".
%The greedy predictions may endup using one of the predicted captions, but an average of distribution of predicted captions will be better representation of baseline caption from the model.
%We propose a variant of SCST where the “baseline” score is obtained using a Bayesian DNN model, which infers the distribution of these predicted captions over model parameters. We use MC dropout to infer and sample from this distribution. The average score of the captions sampled from this distribution will be a better representation of the “baseline” score.
%Distribution of output captions are generated using a Bayesian DNN model and the average of their scores for the generated captions is used as the “baseline” score, which indicates predictive confidence of the model. 
%The expected reward using this “baseline” score is back-propagated to boost words in the captions with higher scores than the baseline while suppressing the words from the captions with lower scores.
%We refer to this approach as Bayesian SCST (B-SCST).
%We demonstrate that B-SCST applied to CIDEr-D scores improves the captioning quality scores, as compared to SCST approach. We also observe that it improves uncertainty measures of the generated captions, as compared to SCST approach.
%\end{comment}
We instead propose a “baseline” that is obtained using a Bayesian DNN model based on MC dropout approximate inference, which infers the distribution of these predicted captions over model parameters. In Figure~\ref{fig:example}~(b), we show the distribution of CIDEr-D scores for 50 randomly selected images from Flickr30k dataset obtained using MC dropout forward passes. For each image, we plot range of CIDEr-D scores around their predictive mean (shown in red).
%We use MC dropout approximate inference to infer this distribution and sample from it.
The predictive mean CIDEr-D score of the captions that are sampled from this distribution will be a better representation of the “baseline”.
%The expected reward using this “baseline” score is back-propagated to boost words in the captions with higher scores than the baseline while suppressing the words from the captions with lower scores.
We refer to this approach as Bayesian SCST (B-SCST).

\noindent In summary, our main contributions in this work are:
\vspace{-4pt}
\begin{itemize}
\item We propose B-SCST, a Bayesian variant of SCST approach, and demonstrate that it improves the caption quality score compared to SCST approach.
\item We present uncertainty quantification of the model generated image captions and demonstrate a good correlation between CIDEr-D scores and uncertainty. To our knowledge, this is the first work which provides this kind of Bayesian analysis for image captioning, and can improve interpretability of generated captions.
\end{itemize}
\vspace{-4pt}

The paper is organized as follows. The related work is presented in Section~\ref{sec:rel_work}, followed by proposed method in Section~\ref{sec:B-SCST}. The results are presented in Section~\ref{sec:results} and conclusions in Section~\ref{sec:conclusions}.

\vspace{-8pt}
\section{Related work}
\label{sec:rel_work}
%In this section, we provide a brief review of relevant work on imaging captioning approaches that provide state-of-the-art results.
\vspace{-10pt}
\subsection{Attention Mechanism}
\vspace{-8pt}
%The recent state-of-the-art approaches for image captioning models use attention mechanism in encoder to generate the latent features used by the decoder network.
State-of-the-art image captioning DNNs use attention mechanism~\cite{xu2015show} so that the encoder and decoder in the model attend on appropriate features in the image to generate the words in the caption. A bottom-up attention mechanism (Up-Down) was proposed in~\cite{anderson2018bottom}, where the model attends on Faster R-CNN~\cite{ren2015faster} proposals from the image.
%Transformer networks\cite{vaswani2017attention}, which are based on self attention, achieved state-of-the-art results on machine translation tasks.
Attention-on-attention network (AoANet)~\cite{huang2019attention} uses an extra learned attention on top of self-attention~\cite{vaswani2017attention} to avoid attentions that are irrelevant to the decoder. In this work, we demonstrate our approach on AoANet model architecture, although it can be applied on other architectures that benefit from SCST.

\vspace{-8pt}
\subsection{Training techniques}
\vspace{-8pt}
Image captioning DNNs are usually trained with word level cross-entropy loss between the ground truth and model generated caption. SCST~\cite{rennie2017self} uses policy-gradient based RL to directly optimize the non-differentiable Natural language processing (NLP) metrics that are used for caption evaluation. Specifically, it uses the caption score obtained by the model using its inference algorithm, as the “baseline” to
%normalize the caption scores generated during its search process and therefore
reduce the variance of gradients.
%A variant of this approach~\cite{anderson2018bottom} uses beam search and samples only from the top-k words at each time step and hence top-k obtained using beam search while keeping the greedy inference score as the “baseline”. Another variant of this method ~\cite{cornia2019m}, where the “baseline” is changed to use the mean of top-k caption scores.  
A few variants of SCST have been used to improve the image captioning quality metrics. One variant~\cite{anderson2018bottom} performs beam search and restricts the search space to only the top-k captions in the decoded beam. Another variant~\cite{cornia2019m} similarly restricts the search space to only the top-k captions in the decoded beam, while also using the mean CIDEr-D score of these top-k captions as the “baseline”. We discuss the differences between our approach and these works in Section 3.

\vspace{-8pt}
\subsection{Bayesian approaches}
\vspace{-8pt}
%Deep neural networks are prone to overfitting and can often make overly confident predictions, which makes them incapable of assessing uncertainty \cite{blundell2015weight}. Bayesian models, on the other hand, treat the parameters of the model as random variables and the output of the model depends on the parameter drawn from their posterior distribution. So, the variance in this output can help in estimating uncertainty measures in the model.  \cite{gal2016dropout} casts dropout as a bayesian approximation which avoids computational cost of bayesian models, but still helps provide uncertainty measures. \cite{fortunato2017bayesian} uses LSTM trained using bayesian backpropagation \cite{blundell2015weight} instead of traditional backpropagation to improve the perplexity of image captioning.  \cite{hama2019exploring} explore uncertainty measures to improve caption embedding retrieval performance. \cite{xiao2019quantifying} use MC dropout along with explicit outputs that predict the model uncertainty. Our approach is different from these works as in we focus on improving image captioning metrics by using MC dropout to cast the state of art model as bayesian, without making any architecture changes to them.

Standard DNNs do not capture uncertainty estimates~\cite{blundell2015weight} associated with the data and the model parameters. SoftMax probabilities obtained from DNNs can often provide overly confident results for incorrect predictions. Hence, Bayesian DNNs have been proposed to capture data and model uncertainties~\cite{blundell2015weight} resulting in more robust models. In~\cite{gal2016dropout}, dropout training in DNNs is cast as an approximate Bayesian inference in deep Gaussian processes. This work has been extensively used in many applications~\cite{kendall2017multi,kendall2015bayesian} to model uncertainty estimates.
Among image captioning related works, an LSTM trained using Bayesian back-propagation was proposed in~\cite{fortunato2017bayesian} to improve the perplexity of image captioning results. Uncertainty measures were explored in~\cite{hama2019exploring} to improve caption embedding and retrieval task. In~\cite{xiao2019quantifying}, MC dropout was used along with explicit outputs that predict the model uncertainty. Our approach is different from these works. We focus on improving image captioning metrics by using MC dropout to cast a state-of-the-art model as Bayesian DNN, without making any changes to the model architecture. We also perform uncertainty quantification of the generated caption to improve their interpretability, which to our knowledge has not been done in earlier works. %has not been done earlier to the best of our knowledge.

\begin{figure}[t]
\centering
\includegraphics[width=0.99\linewidth]{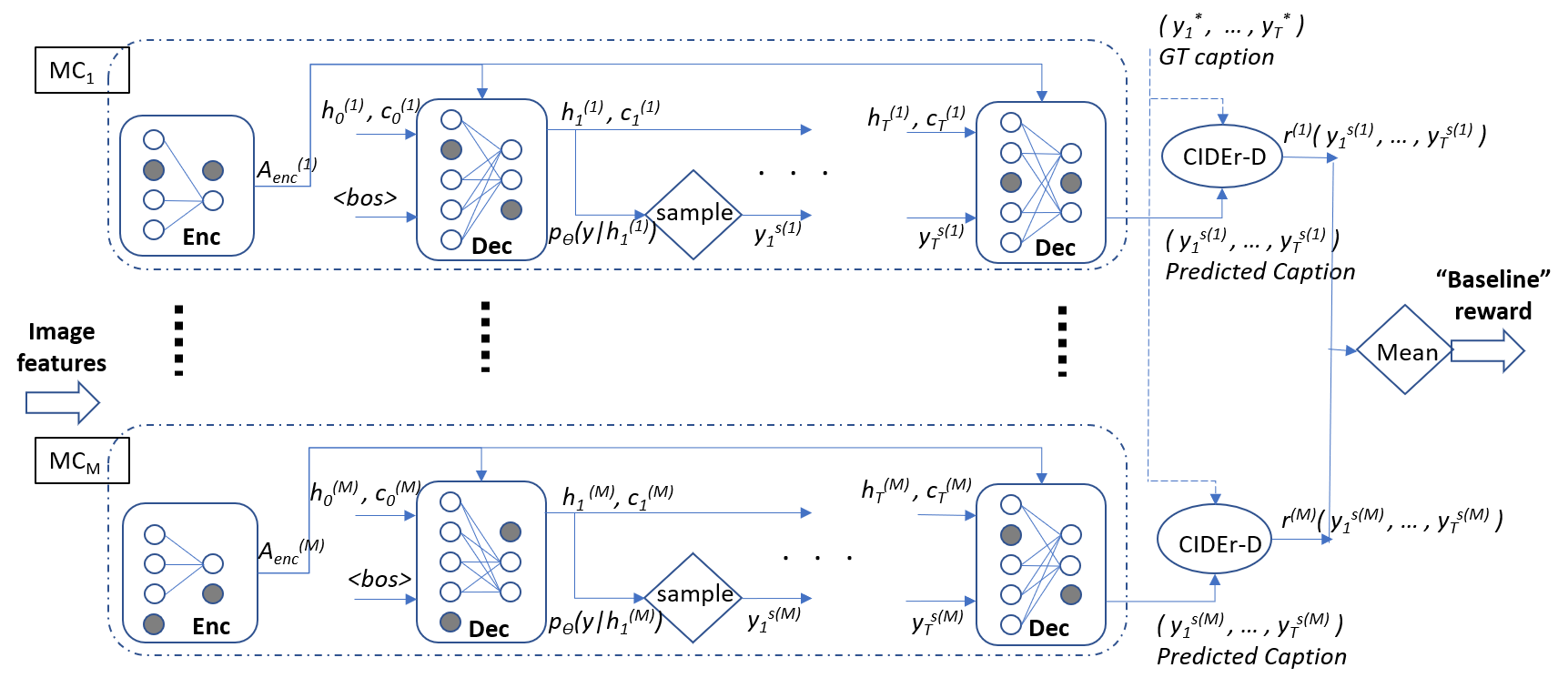}
\caption{\small Bayesian Self Critical Sequence Training (B-SCST). The AoANet encoder and decoder are marked as “Enc” and “Dec” respectively, and the gray nodes inside them indicate the dropout nodes. The $M$ MC dropout forward passes through the model are marked as $MC_1$ through $MC_M$. In each of these forward passes, the input image features go through the encoder, and the decoder uses them to generate the word prediction at each time-step. CIDEr-D score is calculated between the predicted caption and the ground truth caption. The predictive mean of the CIDEr-D scores from these forward passes is used as the “baseline” during policy gradient RL training.}
\label{fig:b-scst}
\end{figure}
%We present our approach in the next section followed by the results in Section~\ref{sec:results}.

\section{Bayesian Self-Critical Sequence Training}
\label{sec:B-SCST}
\vspace{-8pt}
In this section, we present details of our approach after discussing the relevant background.
\vspace{-4pt}
\subsection{Background}
\vspace{-8pt}
\subsubsection{Model Architecture}
\vspace{-6pt}
We used AoANet architecture~\cite{huang2019attention} for our trials since it recently provided state-of-the-art results. AoANet is based on encoder-decoder architecture with attention mechanism.
Given an image $I$, Faster R-CNN is used to extract
%set of $k$
feature vectors~$A$.
%= $\{a_1,a_2,…,a_k\}$, where $a_i \in \mathbb{R}^D$, and D is dimension of each vector.
The AoANet encoder, which includes AoA modules, generates re-weighted feature vectors~$A_{enc}$. At each timestep~$t$ (i.e., word in the caption), the AoANet decoder, which includes Attention LSTM and its own AoA modules, uses~$A_{enc}$ and previous word~$y_{t-1}$ in the caption to generate hidden state $h_t$ and context vector~$c_t$. The context vector~$c_t$ is used to compute the conditional probabilities of the words in the word vocabulary $p_{\theta}(y_t\,|\,y_{1:t-1}, I)$, where~$\theta$ are the AoANet model parameters. Additional AoANet architecture details can be found in~\cite{huang2019attention}.

%To simplify notation in the rest of this paper, we denote the model generated caption of maximum length~$T$ as~$y_{1:T}$ and conditional probability vector of its words as~$p_{\theta}(y_{1:T})$.
To simplify notation in the rest of this paper, we denote the model generated caption as~$y_{1:T}$, and conditional probability vector of its words as~$p_{\theta}(y_{1:T})$, where~$T$ is the maximum length of the caption.

\vspace{-8pt}
\subsubsection{Bayesian Inference}
\vspace{-8pt}
In Bayesian DNNs, model parameters $\theta$ are treated as random variables with a prior distribution $p(\theta)$, instead of point estimates. Given input-output pairs $(x,y)$ and the model likelihood $p(y|x,\theta)$, Bayes rule can be used to obtain the posterior distribution of model parameters $p(\theta|x,y)$:
\begin{equation}
p(\theta|x,y) = \frac{p(y|x,\theta)p(\theta)}{p(y|x)}
\label{eq:post_dist}
\end{equation}
%Since computing this posterior distribution $p(\theta|x,y)$ is often intractable, Bayesian approximate inference techniques such as~(i)~Markov Chain Monte Carlo (MCMC) sampling based probabilistic inference~\cite{neal2012bayesian}, (ii)~Variational inference techniques~\cite{graves2011practical,blundell2015weight}, and (iii)~MC dropout approximate inference~\cite{gal2016dropout}, are often used to infer the tractable approximate posterior distribution $q_\phi(\theta)$.
Since computing this posterior distribution $p(\theta|x,y)$ is often intractable, Bayesian approximate inference techniques~\cite{neal2012bayesian,graves2011practical,blundell2015weight,gal2016dropout} are used to infer a tractable approximate posterior distribution~$q_\phi(\theta)$.
Given a new input $x^*$, the predictive distribution of the output $y^*$ during inference phase is obtained using multiple stochastic forward passes through the model, while sampling from the approximate posterior distribution of model parameters~$q_\phi(\theta)$ using Monte Carlo estimators:  %Equation~\ref{eq:pred_dist} shows the predictive distribution of the output $y^*$ given new input $x^*$:
\begin{equation}
\begin{gathered}
%p(y^*|x^*,x,y) = \int p(y^*|x^*,\theta)\,q_\phi(\theta) d\theta \\
p(y^*|x^*,x,y) \approx \frac{1}{M} \sum_{m=1}^{M}p(y^*|x^*,{\theta}_m)~,~~~{\theta}_m\sim q_\phi(\theta)
\end{gathered}
\label{eq:pred_dist}
\end{equation}
where, $M$ is number of Monte Carlo samples.
%For image captioning model in Section\ref{sec:Model Architecture}, this translates to:
%\begin{equation}
%p_{\theta}(y_{t}|y_{1:t-1},I) \approx \frac{1}{M} %\sum_{i=1}^{M}p_{\theta}(y_{t}|y_{1:t-1},I,{\theta}_i)~,~~~{\theta}_i\sim q_\phi(\theta)
%\label{eq:pred_dist_imgcap}
%\end{equation}

In this work, we use MC dropout approximate inference to obtain the "baseline" for RL step during training phase, and also to perform uncertainty quantification analysis during inference phase. More details are presented in the next two sections.
%We denote $p_{\theta}(y_t\,|\,y_{1:t-1},I)$ as $p_{\theta}(y_t\,|\,y_{1:t-1})$ in the rest of paper to simplify the notations.
\vspace{-4pt}
\subsection{Our Approach}
\vspace{-8pt}
\subsubsection{Training}
\vspace{-8pt}
We train our image captioning model using the popular two step approach.
%In the first step, the cross-entropy loss is used as the objective function, and then followed policy-gradient based reinforcement learning that optimizes non-differentiable metrics such as CIDEr-D scores.
In the first step, we minimize word-level cross-entropy loss function \cite{anderson2018bottom,huang2019attention}.
%~~~~~~~~~~~~~~~~~~~~~~~~~~~~~~~~~~~~~~~~~~~~~~~~~~~~~~~~~~~~~~~~~~~~~~~~~~~~~~~~~~~~~~~~~~~~~~~~~~~~~~~~~~~~~~~~~~~~~
%\noindent\textbf{Cross-entropy loss optimization}: We first minimize word-level cross-entropy loss function \cite{anderson2018bottom,huang2019attention}:
%\begin{equation}
%L_{XE}(\theta) = -\sum_{t=1}^T %\log(p_{\theta}(y_t^*~|~y_{1:t-1}^*))
%\end{equation}
%where, $y_{1:T}^*$ is the ground truth caption, $\theta$ are the model parameters, $T$ is number of words in the caption and $p(y_t\,|\,y_{1:t-1})$ is the conditional probability of the generated word in the vocabulary at timestep $t$.
%~~~~~~~~~~~~~~~~~~~~~~~~~~~~~~~~~~~~~~~~~~~~~~~~~~~~~~~~~~~~~~~~~~~~~~~~~~~~~~~~~~~~~~~~~~~~~~~~~~~~~~~~~~~~~~~~~~~~~
%\noindent\textbf{CIDEr-D optimization}:
In the second step, we optimize CIDEr-D directly and refer to this step as CIDEr-D optimization. In this work, we use our proposed B-SCST for CIDEr-D optimization. We first briefly describe the well-known SCST approach~\cite{rennie2017self}, which works as follows. The decoder (agent) interacts with the image features and the current word in the caption (environment) using the model’s parameters~$\theta$ (policy~$p_{\theta}$), to generate the next word (action) in the caption. After the complete caption is generated, CIDEr-D score (reward) is calculated using the ground truth sentence. The goal of CIDEr-D optimization is to minimize the negative expected CIDEr-D rewards function $r(.)$:
\begin{equation}
L_{RL}(\theta) = -\mathbf{E}_{y_{1:T}\sim p_{\theta}}{\left [r(y_{1:T})  \right ]}
\end{equation}
The gradient of this loss is approximated \cite{rennie2017self} as: % using
\begin{equation}
\nabla_{\theta}L_{RL}(\theta) \approx -(r(y_{1:T}^s) - r(\hat{y}_{1:T}))\nabla_{\theta}\log p_{\theta}(y_{1:T}^s)
\label{eq:pol_grad}
\end{equation}
Here,~$y_{1:T}^s$ is the sampled caption that is generated by sampling words from the decoder's output SoftMax probability distribution over the word vocabulary at each timestep.~$\hat{y}_{1:T}$ is the greedy caption that is generated using inference algorithm, i.e., by choosing the word with highest SoftMax probability at each time step. SCST approach uses reward $r(\hat{y}_{1:T})$ of the caption ~$\hat{y}_{1:T}$ as the “baseline” to normalize the rewards of sampled caption~\cite{rennie2017self}, and reduces the variance of the gradient. This gradient formulation increases the probability of the captions with CIDEr-D scores higher than those generated by the current model at inference phase, and decreases the probability of the captions with lower CIDEr-D scores.

The choice of “baseline” is important here, and the usage of CIDEr-D score of a single greedy caption as “baseline” may be undesirable when there is uncertainty in the model predictions.
%As shown in Figure~\ref{fig:example},  there could be uncertainty in greedy caption and its corresponding CIDEr-D reward score, which could lead to underestimating or overestimating the “baseline”.
%The SoftMax probabilities have been shown to be overly confident~\cite{gal2016dropout}, i.e., possibly high values for incorrect predictions, even when the model is uncertain about its predictions.
%As shown in Figure~\ref{fig:example} and also later in Section~\ref{sec:results} (Figure~\ref{fig:uncert-vs-cider-flickr}), there could be uncertainty in greedy caption and its corresponding CIDEr-D reward score, which could lead to underestimating or overestimating the “baseline”.
We, therefore, propose to use Bayesian inference to estimate the “baseline”, and refer to it as B-SCST.
In B-SCST, we use MC dropout approximate inference, i.e., we run multiple MC dropout forward passes through the model, to infer the posterior distribution of captions around the model parameters, and estimate their predictive mean CIDEr-D score (example is shown in Figure~\ref{fig:example}~(b)). The dropout layers are modeled using Bernoulli distribution~\cite{gal2015bayesian} with dropout rate as the parameter.
%We use multiple MC dropout forward passes through the model to infer the distribution of captions around the current model parameters, and estimate their mean CIDEr-D score.
We use this predictive mean CIDEr-D score, which accounts for the uncertainty, as the “baseline” during CIDEr-D optimization.
%Also, we do not perform any greedy sampling, i.e., choosing only the word with highest SoftMax probability, during this process. Instead, we sample from the SoftMax distribution during training, which allows the model to explore a larger search space. We observe improved results due to this in our ablation study shown in Section~\ref{sec:ablation}. % when words with lower SoftMax are sampled.

During each MC dropout forward pass of training phase, we sample words from the decoder's output SoftMax probability distribution in order to generate a caption. We do not choose the word(s) with highest SoftMax probability here, and instead sample from the SoftMax distribution, in order to allow the model to explore a larger search space. We observe improved results with our sampling approach, as shown later in our ablation study in Section~\ref{sec:ablation}.
The "baseline" $\tilde{r}$ and gradient of the loss in our proposed model can be approximated by changing Equation~\ref{eq:pol_grad} as:
%The predictive confidence scores obtained by considering mean of the predictive distribution are used as the “baseline” score. We do not perform any “greedy” sampling during this step allowing the model to explore a larger search to calculate the rewards values. The gradient of loss in our proposed model can be approximated as:
\begin{equation}
\begin{gathered}
\tilde{r} \approx \frac{1}{M}\sum_{m=1}^Mr({{y}_{1:T}^s}^{(m)}) \\
%\,\,\,\,\,\text{and}\,\,\,\,
\nabla_{\theta}L_{RL}(\theta) \approx -\frac{1}{M}\sum_{m=1}^M(r({y_{1:T}^s}^{(m)}) - \tilde{r})\nabla_{\theta}\log p_{\theta}({y_{1:T}^s}^{(m)})
%r(\tilde{y}_{1:T}) = \frac{1}{M}\sum_{m=1}^Mr({{y}_{1:T}^s}^{(m)}) \\
%\nabla_{\theta}L_{RL}(\theta) \approx -\frac{1}{M}\sum_{m=1}^M(r({y_{1:T}^s}^{(m)}) - r(\tilde{y}_{1:T}))\nabla_{\theta}\log p_{\theta}({y_{1:T}^s}^{(m)})
%\nabla_{\theta}L_{RL}(\theta) \approx -\frac{1}{M}\sum_{m=1}^M(r({y_{1:T}^s}^{(m)}) - \frac{1}{M}\sum_{n=1}^Mr({{y}_{1:T}^s}^{(n)}))\nabla_{\theta}\log p_{\theta}({y_{1:T}^s}^{(m)})
\end{gathered}
\end{equation}
where $M$ is the total number of MC dropout forward passes, and ${y_{1:T}^s}^{(m)}$ is the sampled caption that is generated during the $m^{th}$ forward pass. We illustrate B-SCST approach in Figure~\ref{fig:b-scst}.
While performing MC dropout during training phase, we enable dropout in both the encoder and decoder in order to capture the model uncertainty. We allow different dropout masks across timesteps of the decoder, since we observed better captioning metric results compared to using the same dropout mask across all decoder timesteps as proposed in~\cite{gal2016theoretically}.

%We want to point out that our approach is different from some of the other works~\cite{cornia2019m,anderson2018bottom}. In~\cite{cornia2019m}, beam-search is used to select the top-5 captions in terms of SoftMax probability, and the mean of five caption scores in the decoded beam is used as the “baseline” reward.
%We do not use beam-search during training, and instead use Bayesian inference by performing multiple stochastic MC dropout forward passes through the model. The captions sampled from beam search would all have the same dropout mask, which is not the case with our approach. We also don't restrict the search space to top-5 candidates.
%We do not use beam-search during training. Beam-search does not involve multiple stochastic MC dropout forward passes through the network like our approach, which encourages Bayesian inference. Also, we do not restrict our sampling search space to only the top-5 SoftMax probability captions, and instead allow lower SoftMax probability captions to be sampled based on their SoftMax distribution.
%Our approach is also different from~\cite{anderson2018bottom}, who also use beam-search to restrict the search space of sampled captions, similar to~\cite{cornia2019m}, but directly use the greedy caption to estimate “baseline” reward, similar to SCST.

We want to point out that our approach is different from some of the other works~\cite{anderson2018bottom,cornia2019m}.
In~\cite{anderson2018bottom}, beam-search is used to to restrict the search space to top-k captions in terms of SoftMax probability, and the greedy caption is used to estimate “baseline”, similar to SCST.
Similary in~\cite{cornia2019m}, beam-search is used to select the top-5 captions in terms of SoftMax probability, but the mean of five caption scores in the decoded beam is used as the “baseline”.
We do not use beam-search during training phase, and instead use Bayesian inference by performing multiple stochastic MC dropout forward passes through the model. The captions sampled from beam search would all have the same dropout mask, which is not the case with our approach. We also don't restrict the search space to top-5 candidates, as we mentioned earlier.
%We do not use beam-search during training. Beam-search does not involve multiple stochastic MC dropout forward passes through the network like our approach, which encourages Bayesian inference. Also, we do not restrict our sampling search space to only the top-5 SoftMax probability captions, and instead allow lower SoftMax probability captions to be sampled based on their SoftMax distribution.
%Our approach is also different from~\cite{anderson2018bottom}, who also use beam-search to restrict the search space of sampled captions, similar to~\cite{cornia2019m}, but directly use the greedy caption to estimate “baseline” reward, similar to SCST.

%, and instead restricts the search space to top-5 captions. Our ablation studies have indicated this can be restrictive in cases where more exploration is needed. We do not restrict our sampling search space to top-5 captions but allow all the samples from the MC simulations. Our approach is also different from ~\cite{anderson2018bottom} who also use beam-search, but directly use the greedy captions to estimate “baseline” reward.

\vspace{-8pt}
\subsubsection{Uncertainty Quantification}
\label{sec:uncert-metric}
\vspace{-4pt}
Bayesian modeling allows capturing estimates for both aleatoric uncertainty, i.e. noise inherent in input observations, and epistemic uncertainty, i.e uncertainty related to model parameters~\cite{gal2016uncertainty, smith2018understanding}. Predictive entropy is a measure of both aleatoric and epistemic uncertainties, whereas mutual information (MI) between model parameters (posterior distribution) and data (predictive distribution) %, also referred to as Bayesian active learning by disagreement (BALD)~\cite{houlsby2011bayesian},
is a measure of epistemic uncertainty~\cite{smith2018understanding}.

%“Aleatoric” or input uncertainty captures noise inherent in the input observations, where as “Epistemic” or model uncertainty captures uncertainty related to model parameters. In this study, we evaluate the model uncertainty using Bayesian active learning by disagreement (BALD) \cite{houlsby2011bayesian}, which quantifies mutual information between parameter posterior distribution and predictive distribution.
%BALD captures “model uncertainty”, and is calculated using Equation~\ref{eq:mutual information}.
%\begin{equation}
%BALD := H(y^*|x^*, x,y)-\E_{p(w|x,y)}[H(y^*|x^*, w)]\\
%\label{eq:mutual information}
%\end{equation}
%where the first term $H(y^*|x^*, x,y)$ is the predictive entropy that captures a combination of both input and model uncertainty, and is calculated as:
% Equation~\ref{eq:pred_entropy}.
%\begin{equation}
%H(y^*|x^*, x,y):=-\sum_{i=0}^{K-1}p_{i\mu} * log\,p_{i\mu}\\
%\label{eq:pred_entropy}
%\end{equation}
%where, $K$ is total number of output classes and $p_{i\mu}$ is the predictive mean SoftMax probability of $i^{th}$ class from $M$ MC samples (given by Equation~\ref{eq:pred_dist}). The second term $\E_{p(w|x,y)}[H(y^*|x^*, w)]$ in~\ref{eq:mutual information} is the mean entropy, where~$H(y^*|x^*, w)$ is the entropy of each MC sample. In the case of image captioning, $K$ is the vocabulary size of the dataset.

In order to estimate uncertainty and perform Bayesian analysis during inference phase, we use MC dropout approximate inference by enabling dropout in the final fully connected layer, and greedily generate the caption with highest SoftMax probability. For a given image, we perform~$M$ MC dropout forward passes to generate~$M$ captions. For each word (i.e., timestep~$t$) in the caption of length $T_m$, that is generated during the~$m^{th}$ MC dropout forward pass, we obtain the SoftMax probability of each class~$i$ in the word vocabulary~$V$, denoted as~${v_t}^{i(m)}$.
%We concatenate SoftMax probabilities of all the words in the caption to get the SoftMax probability of the entire caption, denoted as~${v_{1:T}}^w$.
We calculate entropy~$H_m$ of the caption that is generated during of~$m^{th}$ MC dropout forward pass, and the mean entropy~$\bar{H}$ of all the~$M$ captions using:
\begin{equation}
\begin{gathered}
H_m \approx \frac{1}{T_m*V}\sum_{t=1}^{T_m}\sum_{i=1}^{V}{v_t}^{i(m)} * log\,{{v_t}^{i(m)}}\,\,\,\,\,\,\text{and}\,\,\,\,\,\,\,
\bar{H} \approx \frac{1}{M}\sum_{m=1}^{M}{H_m}.\,\,\,\,\,\,\,\,\,
\label{eq:entropy}
\end{gathered}
\end{equation}
%We give a simplified equation here, since caption lengths can vary across MC simulations. We use Python numpy package "nan" functions appropriately to account for this.
%The mean entropy of the MC dropout captions is then given by:
%\begin{equation}
%\bar{H} \approx \frac{1}{M}\sum_{m=1}^{M}{H_m}\\
%\label{eq:mean_entropy}
%\end{equation}
We calculate the predictive entropy of these MC dropout captions using:
\begin{equation}
\begin{gathered}
H \approx \frac{1}{T*V}\sum_{t=1}^{T}\sum_{i=1}^{V}{\bar{v_t}}^i * log\,{\bar{v_t}^i}\,\,\,\,\,\,\,\text{where}\,\,\,{\bar{v_t}}^i = \frac{1}{M}\sum_{m=1}^{M}{v_t}^{i(m)}\,\,\,\,\,\,
\label{eq:pred_entropy}
\end{gathered}
\end{equation}

Mutual information (MI) is then given by:~$MI := H -\bar{H}$
%\begin{equation}
%BALD := H -\bar{H}\\
%\label{eq:mutual information}
%\end{equation}
%For each MC sample, we obtain the SoftMax probability of the generate image caption by concatenating the SoftMax probabilities of all the words in the caption, where each word's Softmax probability contains all the output classes (i.e., vocabulary).
%The predictive distribution of the captions, obtained using 30 MC dropout forward passes, is used to estimate BALD (Equation~\ref{eq:mutual information}) and predictive entropy (Equation~\ref{eq:pred_entropy}) uncertainty scores.

\begin{table}[!tb]
\small
\renewcommand{\arraystretch}{1.3}
\begin{center}
\begin{tabular}{M{0.55 cm}|M{1.35 cm}|C{0.325 cm} C{0.35 cm} C{0.325 cm} C{0.325 cm} C{0.6 cm}C{0.25 cm} C{0.001cm} | M{0.325 cm} M{0.35 cm} M{0.325 cm} M{0.325 cm} M{0.6 cm} M{0.4 cm}}
&  & \multicolumn{6}{c}{Cross-entropy loss training} && \multicolumn{6}{c}{CIDEr-D optimization}   \\ \cline{3-15}
& Model & B@1 & B@4 & M & R & C & S && B@1 & B@4 & M & R & C & S \\ \hline\hline
\multicolumn{2}{c}{} & \multicolumn{13}{c}{Flickr30k} \\\hline%\cline{3-15}
Test & SCST & 69.6 & 28.0 & 22.2 & 48.8 & 58.5 & 16.4 && 72.2 & 30.0 & 22.1 & 50.0 & 64.6 & 16.3 \\
Split& B-SCST & 69.6 & 28.0 & 22.2 & 48.8 & 58.5 & 16.4 && 71.9 & 29.6 & 22.6 & 50.2 & \textbf{66.9} & 16.7 \\\hline
% & Metric & B@1 & B@4 & M & R & C & S & B@1 & B@4 & M & R & C & S \\
Val& SCST & 69.5 & 27.8 & 22.1 & 49.1 & 59.9 & 16.0 && 72.6 & 29.9 & 22.0 & 50.3 & 64.5 & 15.7 \\
Split & B-SCST & 69.5 & 27.8 & 22.1 & 49.1 & 59.9 & 16.0 && 72.4 & 29.1 & 22.5 & 50.3 & \textbf{67.0} & 16.2 \\\hline\hline
%------------------------------------------------------
\multicolumn{2}{c}{} & \multicolumn{13}{c}{MS COCO} \\\hline%\cline{3-15}
\multirow{3}{0.75 cm}{Test Split} & SCST*~\cite{huang2019attention} & 77.4 & 37.2 & 28.4 & 57.5 & 119.8 & 21.3 && 80.2 & 38.9 & 29.2 & 58.8 & 129.8 & 22.4 \\
& SCST & 77.3 & 36.9 & 28.5 & 57.3 & 118.4 & 21.7 && 80.5 & 39.1 & 29.0 & 58.9 & 128.9 & 22.7  \\
& 	B-SCST & 77.3 & 36.9 & 28.5 & 57.3 & 118.4 & 21.7 && 80.8 & 39.0 & 29.2 & 59.0 & \textbf{131.0} & 22.9 \\\hline
%\multirow{3}{0.75 cm}{Val Split}  & SCST*~\cite{huang2019attention} & N/A & N/A & N/A & N/A & N/A & N/A && N/A & N/A & N/A & N/A & N/A & N/A \\
\multirow{2}{0.75 cm}{Val Split} & SCST & 77.3 & 37.3 & 28.3 & 57.4 & 117.4 & 21.4 && 80.4 & 39.1 & 28.9 & 58.9 & 127.7 & 22.5 \\ %\\\hline\hline
& 	B-SCST  & 77.3 & 37.3 & 28.3 & 57.4 & 117.4 & 21.4 && 80.8 & 39.0 & 29.0 & 58.9 & \textbf{129.4} & 22.7 \\ \hline\hline
%------------------------------------------------------
\multicolumn{2}{c}{} &  \multicolumn{13}{c}{VizWiz} \\\hline%\cline{3-15}
Test& SCST*~\cite{gurari2020captioning} & - & - & - & - & - & - &&  66.0 & 23.7 & 20.1 & 46.8 & 60.9 & 15.3  \\
Split & B-SCST & 64.7 & 22.7 & 19.4 & 45.0 & 59.0 & 14.7 && 66.3 & 24.0 & 20.3 & 46.9 & \textbf{63.7} & 15.7  \\
%------------------------------------------------------
\end{tabular}
\end{center}

\caption{\small Results on Flickr30k~\cite{plummer2015flickr30k}, MS COCO~\cite{chen2015microsoft} and VizWiz~\cite{gurari2020captioning} datasets. Our approach B-SCST consistently improves the CIDEr-D scores as compared to the traditional SCST approach. SCST* scores are presented from the published work. For MS COCO-Test split, we observe SCST scores are slightly lower than SCST* published~\cite{huang2019attention} scores while using the model check-point provided by the authors.}

\vspace*{-8pt}
\label{tab:all_bscst_results}
\end{table}

%\vspace{-4pt}
\section{Experiments}
\label{sec:results}
\vspace{-8pt}
%In this section, we present the image captioning results for B-SCST approach and compare against a model where the SCST approach is used. B-SCST can be applied to other image captioning architectures that are trained using SCST to maximize non-differentiable metric scores, such as CIDEr-D.

\subsection{Datasets}
\vspace{-8pt}
\label{sec:datasets}
We present the image captioning results on Flickr30k~\cite{plummer2015flickr30k}, MS COCO~\cite{chen2015microsoft} and VizWiz~\cite{gurari2020captioning} image captioning datasets. We compare the standard image captioning evaluation metrics, including BLEU~\cite{papineni2002bleu}, METEOR~\cite{denkowski2014meteor}, Rouge-L~\cite{linrouge}, CIDEr-D~\cite{vedantam2015cider} and SPICE~\cite{anderson2016spice} scores.

\noindent\textbf{Flickr30k:} We use Flickr30k data splits from~\cite{karpathy2015deep}, which contain 31014 training, 1014 validation and 1000 test images, each of which have 5 ground truth captions as labels. All captions are converted to lower case~\cite{huang2019attention,lu2018neural} and only the words occuring at least 5 times are used to build a vocabulary of size 7000 words. We use the bottom-up image features from~\cite{zhou2019grounded} as input to our encoder stage.\newline
\noindent\textbf{MS~COCO:} We use MS COCO data splits from~\cite{karpathy2015deep}, which contain 113287 training, 5000 validation and 5000 test images, each with 5 ground truth captions as labels. We perform similar text preprocessing as Flickr30k dataset, but use only the words occuring at least 4 times in order to build a vocabulary of size 10369 that matches  AoANet~\cite{huang2019attention} vocabulary. We use the bottom-up image features from~\cite{anderson2018bottom} that were  generated using Faster R-CNN model pretrained on ImageNet and Visual genome datasets.
\noindent\textbf{VizWiz:} We use VizWiz data splits from~\cite{gurari2020captioning}, which contain 22866 training, 7542 validation and 8000 test images, each having upto 5 ground truth captions as labels. Following the approach in~\cite{gurari2020captioning}, we combine training and validation images to get for 30408 images for training, and use VizWiz 2020 challenge leaderboard~\cite{vizwizworkshop} to evaluate performance on test set. We perform similar text preprocessing as MS COCO dataset to build a vocabulary of size 7279 that matches baseline~\cite{gurari2020captioning} provided by the challenge organizers.%https://github.com/Yinan-Zhao/AoANet\_VizWiz.
%https://github.com/Yinan-Zhao/AoANet\_VizWiz

\vspace{-10pt}
\subsection{Training and Inference}
\vspace{-8pt}
%We first train the network using cross-entropy loss followed by the proposed B-SCST approach to maximize the CIDEr-D scores~(details are in Section~\ref{sec:B-SCST}). We use a minibatch size of 10 images, with every image having five ground-truth captions, and use ADAM~\cite{kingma2014adam} optimizer.
We train the model using a minibatch size of 10 images and ADAM~\cite{kingma2014adam} optimizer.
We first run 25 epochs of cross-entropy loss training, with optimizer learning rate of 2e-4 and decay factor of 0.8 every 3 epochs. The scheduled sampling~\cite{bengio2015scheduled} probability is increased at a rate of 0.05 every 5 epochs along with label smoothing \cite{szegedy2016rethinking}.
Since each image contains 5 ground truth labels, we replicate each image feature 5 times and pass it through the model to calculate cross-entropy loss for each caption.
We run 30 epochs of CIDEr-D optimization with optimizer learning rate of 2e-5 and a reduce on plateau factor of 0.5 when CIDEr-D degrades for more than one epoch. For VizWiz dataset, we run only 25 epochs of CIDEr-D optimization with starting learning of 1e-5 to avoid overfitting. During inference, we use beam search and pick the caption with the highest SoftMax probability. We use a beam size of 2 for fair comparison of our results with published AoANet and VizWiz results.
During training using B-SCST approach, we use~$M$=5 for VizWiz dataset and ~$M$=10 for MS COCO and Flickr30k datasets. During inference for uncertainty quantification (Section~\ref{sec:uncet-quant-results}), we use~$M$=30 MC simulations with no beam search.

\begin{figure}[t]
{
%\captionsetup[subfloat]{labelfont=bf}
{
\begin{subfigure}{0.32\textwidth}
%\subfloat{\small \textbf{Predictive Entropy}}
\centering
\captionsetup{
justification=centering}
\includegraphics[scale=0.35]{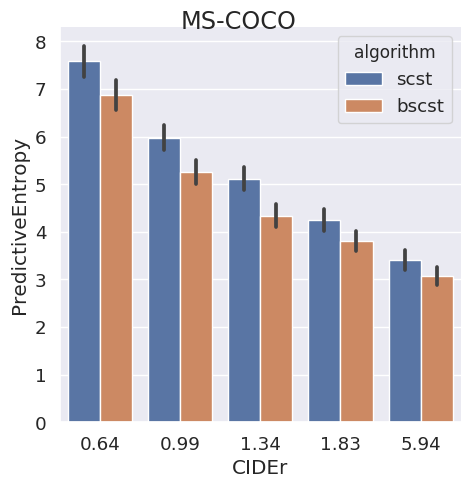}
%\caption{\small Predictive Entropy Uncertainty}
\end{subfigure}
\begin{subfigure}{0.33\textwidth}
%\subfloat{\small \textbf{Mutual Information}}
\centering
\captionsetup{
justification=centering}
\includegraphics[scale=0.35]{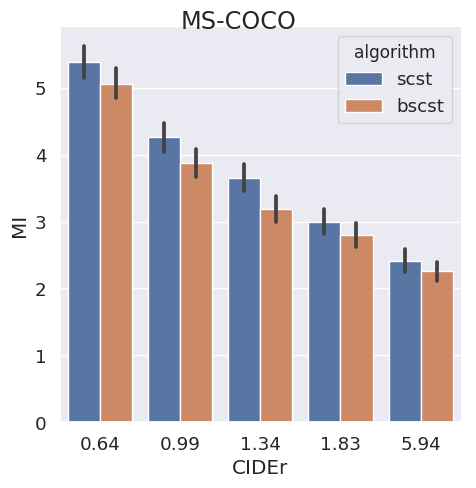}
%\caption{\small BALD Uncertainty}
\end{subfigure}
\begin{subfigure}{0.32\textwidth}
%\subfloat{\small \textbf{SoftMax}}
\centering
\captionsetup{
justification=centering}
\includegraphics[scale=0.35]{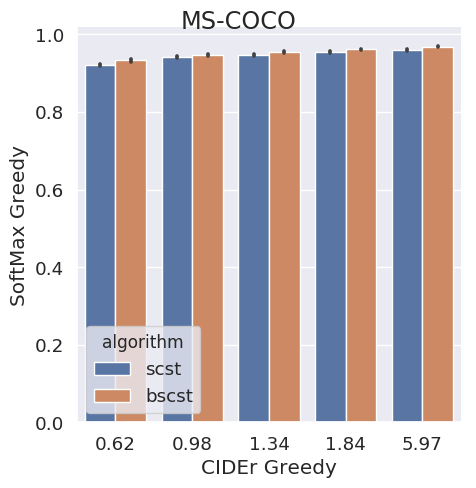}
%\caption{\small SoftMax scores for the caption}
\end{subfigure}
}
%\subcaption{MS-COCO Karpathy val split}

%----------------------------------------------------
{

\begin{subfigure}{0.32\textwidth}
\centering
\captionsetup{
justification=centering}
\includegraphics[scale=0.35]{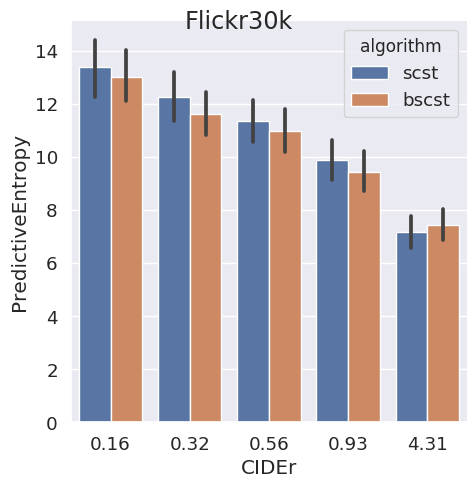}
%\caption{\small Predictive Entropy Uncertainty}
\end{subfigure}
\begin{subfigure}{0.32\textwidth}
\centering
\captionsetup{
justification=centering}
\includegraphics[scale=0.35]{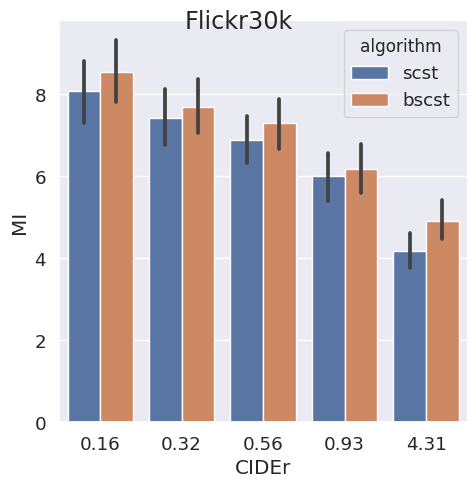}
%\caption{\small BALD Uncertainty}
\end{subfigure}
\begin{subfigure}{0.32\textwidth}
\centering
\captionsetup{
justification=centering}
\includegraphics[scale=0.35]{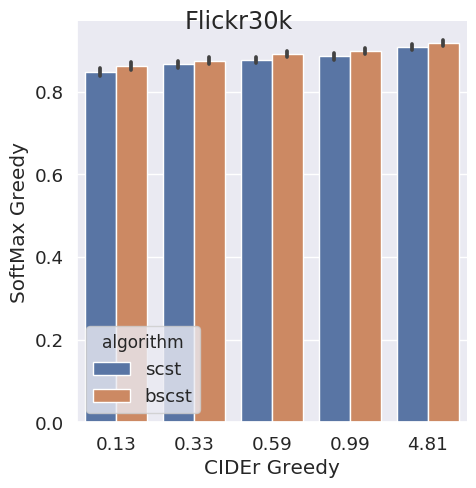}
%\caption{\small SoftMax scores for the caption}
\end{subfigure}
}
%\subcaption{Flikr30k Karpathy val split}
%----------------------------------------------------
{

\begin{subfigure}{0.32\textwidth}
\centering
\captionsetup{
justification=centering}
\includegraphics[scale=0.35]{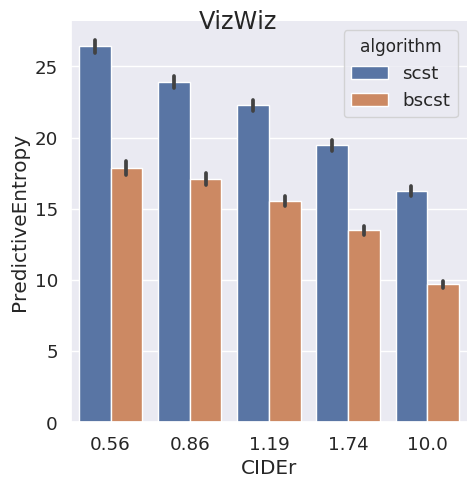}

\end{subfigure}
\begin{subfigure}{0.32\textwidth}
\centering
\captionsetup{
justification=centering}
\includegraphics[scale=0.35]{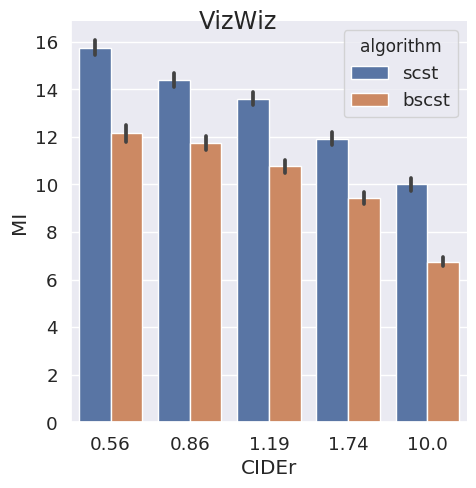}
\end{subfigure}
\begin{subfigure}{0.32\textwidth}
\centering
\captionsetup{
justification=centering}
\includegraphics[scale=0.35]{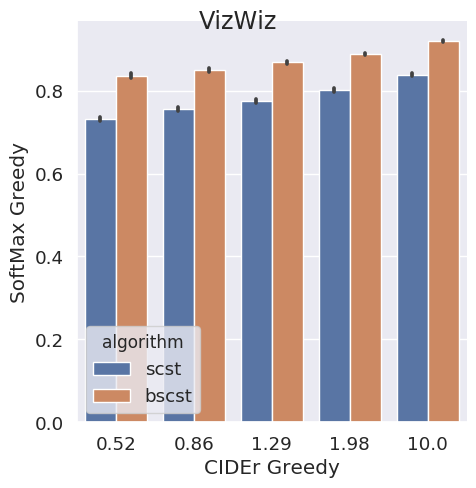}
\end{subfigure}
}
%\subcaption{Vizwiz validation split}
}
\caption{\small
Comparison of Uncertainty vs predictive mean CIDEr-D scores using Bayesian inference (columns 1 \& 2) and SoftMax vs CIDEr-D scores using standard DNN inference (column 3). It demonstrates that the uncertainty estimates (Section~\ref{sec:uncert-metric}) obtained from Bayesian inference are well correlated with the predictive mean CIDEr-D scores, where lower uncertainty (higher confidence) scores are observed for higher CIDEr-D scores. On the contrary, SoftMax probabilities give high scores for different levels of CIDEr-D scores.}
\vspace{-8pt}
\label{fig:uncert-vs-cider-flickr}
\end{figure}

\vspace{-8pt}
\subsection{Image Captioning Results}
\vspace{-8pt}
%We present our results in Tables~\ref{tab:Flickr} and~\ref{tab:MSCOCO}, where B@N, M, R, C and S stand for BLEU@N, METEOR, ROUGE-L, CIDEr-D and SPICE scores, respectively.
In Table~\ref{tab:all_bscst_results}, we present a comparison of SCST and B-SCST approaches for the three datasets considered in our experiments (Section~\ref{sec:datasets}). SCST and B-SCST approaches for CIDEr-D optimization start with the same checkpoint obtained from the cross-entropy loss training.
%In Table~\ref{tab:MSCOCO}, we present the scores on Karpathy test and validation splits of MS COCO dataset.
%On the MS-COCO dataset, we present the scores of AoANet trained with SCST approach  Karpathy test split\footnote{We present val split results in supplemental material.}. results~\cite{huang2019attention}
For MS COCO~\cite{huang2019attention} and VizWiz~\cite{gurari2020captioning} datasets, presented SCST* results are taken directly from the published papers. For MS~COCO dataset, SCST results were obtained using the model checkpoints provided by the authors~\cite{huang2019attention}, which we observe to be slightly lower than the published results. These results in Table~\ref{tab:all_bscst_results} show that B-SCST consistently improves CIDEr-D scores, along with most image captioning metrics, for all the three datasets.

%the AoANet repository\footnote{https://github.com/husthuaan/AoANet}. We found that the model checkpoints shared by the authors provide slightly lower test split scores than the results presented in their paper. In our experiments, we used the cross-entropy loss trained model checkpoint provided by the authors and run CIDEr-D optimization on them to compare SCST and B-SCST methods.
%We observe that B-SCST improves CIDEr-D by 0.8 and 0.9 points on test and val splits, respectively, compared to SCST scores; All the other image captioning quality metric scores are also improved. %In Table~\ref{tab:MSCOCO}, we present the scores on Karpathy validation split of MS COCO dataset.

%These results confirm that the proposed Bayesian inference to obtain “baseline” scores in B-SCST provides improvement in the image captioning quality scores over the SCST approach.
%\textbf{DO WE NEED THIS HERE?} -- Our approach can be extended to other Bayesian models including mean-field variational inference~\cite{hoffman2013stochastic}, where the model parameters are represented by Gaussian distributions.

\vspace{-10pt}
\subsection{Uncertainty Quantification Results}
\label{sec:uncet-quant-results}
\vspace{-8pt}
%We perform MC-dropout during inference by enabling dropout in the final fully connected layer to obtain the MC samples for  Bayesian analysis. The predictive distribution of the captions, obtained using 30 MC dropout forward passes, is used to estimate BALD (Equation~\ref{eq:mutual information}) and predictive entropy (Equation~\ref{eq:pred_entropy}) uncertainty scores.
During inference phase, we perform MC dropout approximate inference (Section~\ref{sec:uncert-metric}) to obtain uncertainty estimates for the captions generated using SCST and B-SCST approaches on the three datasets considered in our experiments~(Section~\ref{sec:datasets}).
In first two columns of Figure~\ref{fig:uncert-vs-cider-flickr}, we plot the uncertainty estimates, i.e., Predictive Entropy and Mutual information (MI) (Section~\ref{sec:uncert-metric}), against predictive mean CIDEr-D scores across MC dropout forward passes.
%\footnote{Plots for MS COCO dataset will be provided in the supplemental material.}
We map CIDEr-D scores into five quantiles
%to obtain bin-edges for the bar plot,
and plot the average uncertainty score for each quantile. We observe that lower CIDEr-D scores indicate higher uncertainty in the predictions, where as higher CIDEr-D scores indicate lower uncertainty. Both these uncertainty measures show good correlation with the CIDEr-D scores, which is critical for the interpretability of the captions generated by the model. We also observe that the uncertainty estimates decrease with the B-SCST approach as compared to SCST for every CIDEr-D quantile, indicating higher confidence in the generated captions. We notice an exception on Flickr30k dataset, where MI is higher with B-SCST approach as compared to SCST, indicating higher model uncertainty although B-SCST results in higher CIDEr-D score (Table~\ref{tab:all_bscst_results}). In practical applications, since the ground truth caption of an image, and therefore its corresponding CIDEr-D score, is not available during inference phase, these uncertainty measures can give an indication of the predictive confidence of the caption generated using the model.  %We also observe that the uncertainty estimates decrease with the B-SCST approach as compared to SCST for every CIDEr-D bin on MS COCO and VizWiz datasets.

In the last column of Figure~\ref{fig:uncert-vs-cider-flickr}, we plot the mean SoftMax probability per word in the caption against the caption's CIDEr-D scores using standard DNN inference, i.e. greedily choosing the word with highest SoftMax at each timestep to generate the caption. We observe that SoftMax probabilities are uniformly distributed for different levels of CIDEr-D scores, further validating that the SoftMax probabilities could be overly confident in predicting the CIDEr-D scores, and not a good measure of predictive confidence of the model.
%The plot for average SoftMax (vs) CIDEr-D scores (Figure\ref{fig:uncert-vs-cider}~(c)) shows SoftMax scores uniformly distributed for different levels of CIDEr-D scores and is not a good measure of predictive confidence of the model.
This uncertainty quantification analysis demonstrates that Bayesian approaches provide robust predictive confidence scores compared to SoftMax probabilities obtained from standard DNN image captioning models. This also justifies the use of a Bayesian "baseline" for the policy-gradient based RL in our B-SCST approach.

\vspace{-10pt}
\subsection{Ablation Studies}
\label{sec:ablation}
\vspace{-8pt}
We perform ablation studies using Flickr30k dataset to confirm the benefits of B-SCST approach and the selection of hyper-parameters. We compare the captioning results on Karpathy validation split with no beam search. In Table~\ref{tab:ablation-studies}~(a), we change the model architecture from AoANet to Up-Down~\cite{anderson2018bottom} and observe that B-SCST approach improves the CIDEr-D score compared to SCST approach. In Table~\ref{tab:ablation-studies}~(b), we use different sampling mechanisms to generate the words in the sampled caption during B-SCST training. These include sampling the word randomly from the word vocabulary (Random), choosing the word with highest SoftMax probability (Top1), sampling the word from the k highest SoftMax probabilities (Topk, with k=5 or 10), and sampling the word from SoftMax probability distribution (Distr.). We observe that sampling approach we use in B-SCST (Distr.) gives the maximum CIDEr-D score. In Table~\ref{tab:ablation-studies}~(c), we vary the number of MC dropout forward passes that are performed during B-SCST training. We observe that increasing number of passes improves the CIDEr-D score with diminishing returns for more than~5\textasciitilde 10 passes. %we vary the number of MC dropout forward passes that are performed to derive the "baseline" during B-SCST training. We observe that "baseline" score from 5\textasciitilde 10 MC dropout samples provides maximum CIDEr-D score improvement.

\begin{table}[t]%[!!htb]
\begin{minipage}{1.0\linewidth}
\centering
\begin{tabular}{M{1.5 cm}|C{0.325 cm} C{0.35 cm} C{0.325 cm} C{0.325 cm} C{0.325 cm}C{0.25 cm} C{0.001cm} | M{0.325 cm} M{0.35 cm} M{0.325 cm} M{0.325 cm} M{0.325 cm} M{0.4 cm}}
& \multicolumn{6}{c}{Cross-entropy loss training} && \multicolumn{6}{c}{CIDEr-D optimization}   \\ \cline{2-14}
Model & B@1 & B@4 & M & R & C & S && B@1 & B@4 & M & R & C & S \\ \hline
SCST & 69.1 & 27.2 & 21.8 & 49.0 & 57.3 & 15.6  && 72.9 & 29.5 & 21.7 & 49.9 & 59.0 & 15.3  \\
B-SCST & 69.1 & 27.2 & 21.8 & 49.0 & 57.3 & 15.6   &&  72.3 & 29.4 & 21.8 & 49.9 & \textbf{61.1} & 15.2 \\
\end{tabular}
\subcaption{Comparison of SCST and B-SCST results using Up-Down model.}
\end{minipage} \hspace{10pt}

\begin{minipage}{.45\linewidth}
\vspace{5pt}
\centering
\begin{tabular}{M{1.2 cm}|C{0.35 cm} C{0.375 cm} C{0.325 cm} C{0.325 cm} C{0.325 cm}C{0.375 cm}}
& B@1 & B@4 & M & R & C & S \\\hline
Random & 68.0 & 25.6 & 21.7 & 48.4 & 58.0 & 15.5 \\\hline
Top1 & 69.3 & 27.3 & 21.6 & 48.5 & 61.6 & 15.3 \\
Top5 & 72.5 & 30.1 & 21.8 & 50.2 & 65.6 & 15.8 \\
Top10 & 72.6 & 29.9 & 21.6 & 50.3 & 64.4 & 15.4 \\\hline
Distr. & 72.4 & 29.6 & 22.6 & 50.5 & \textbf{66.8} & 16.2 \\
\end{tabular}
\subcaption{Comparision of B-SCST sampling approaches.}
\end{minipage}%
\hspace{10pt}
\begin{minipage}{.50\linewidth}
\vspace{5pt}
\centering
\begin{tabular}{M{0.6 cm}|C{0.35 cm} C{0.4 cm} C{0.325 cm} C{0.325 cm} C{0.325 cm}C{0.3 cm}}
\#MC & B@1 & B@4 & M & R & C & S \\\hline
1 & 68.1 & 25.7 & 21.7 & 48.4 & 58.3 & 15.5 \\
3 & 70.7 & 28.3 & 22.3 & 49.8 & 64.2 & 16.1 \\
5 & 72.4 & 29.6 & 22.6 & 50.5 & 66.8 & 16.2 \\
10 & 72.4 & 29.1 & 22.5 & 50.3 & \textbf{67.1} & 16.2 \\
15 & 71.8 & 28.5 & 22.4 & 49.9 & 66.5 & 16.1 \\
\end{tabular}
\subcaption{B-SCST results with different number of MC passes.}
\end{minipage}
\vspace{-8pt}
\caption{B-SCST ablation study results on Flickr30k Karpathy val split.}
\label{tab:ablation-studies}
\end{table}

\vspace{-15pt}
\section{Conclusions}
\label{sec:conclusions}
\vspace{-10pt}
We presented B-SCST for image captioning models, a Bayesian variant of the SCST approach that directly optimizes the CIDEr-D metric.
In B-SCST, we estimate “baseline” for the policy-gradients by averaging CIDEr-D of captions sampled from the distribution inferred using a Bayesian DNN model and demonstrated improved CIDEr-D scores on Flickr30k, MS COCO and VizWiz datasets, as compared to SCST approach.
B-SCST can be applied on other image captioning architectures that benefit from using SCST approach.
%B-SCST method proposed “baseline” reward for the policy-gradients by averaging predictive confidence scores calculated from the distributions of output captions generated by the Bayesian DNN model.
%\textbf{DO WE NEED THIS - CAN THEY BEAT US UP ON IT:)} -- Although the predictive distribution is inferred using MC dropout in our work, we believe the technique can be extended to other Bayesian inference techniques such as mean-field variational inference, where parameters are sampled from Gaussian distributions.
%The proposed approach provides an improvement of in CIDEr-D scores on MS COCO, Flickr30k image captioning datasets, as compared to SCST approach.
We also perform uncertainty quantification analysis on the captions generated using a state-of-the-art captioning model, and demonstrate that these uncertainties correlate well with the CIDEr-D scores and can thus improve interpretability of model generated captions.
\bibliographystyle{unsrt}
\bibliography{bscst_neurips_2020_arxiv}

\end{document}